\documentclass[acmtog, acmlarge]{acmart}

\AtBeginDocument{%
  \providecommand\BibTeX{{%
    \normalfont B\kern-0.5em{\scshape i\kern-0.25em b}\kern-0.8em\TeX}}}

\copyrightyear{2024}

\usepackage{booktabs}
\def \hfillx {\hspace*{-\textwidth} \hfill}
\begin{document}

\title{Towards an educational tool for supporting neonatologists in the delivery room}



\author{Giorgio Leonardi}
\authornotemark[1]
\orcid{0000-0002-9533-9722}
\email{giorgio.leonardi@uniupo.it}
\affiliation{%
  \institution{Department of Science, Technology and Innovation (DiSIT), Computer Science Institute - Università del Piemonte Orientale, Viale Teresa Michel 11, 15121 Alessandria, Italy}
  \streetaddress{Viale Teresa Michel, 11}
  \streetaddress{Viale Teresa Michel, 11}
  \city{Alessandria}
  \country{Italy}
  }

\author{Clara Maldarizzi}
\authornotemark[1]
\affiliation{\institution{Neonatal Intensive Care Unit, Cardinal Massaia Children's Hospital of Asti}\country{Italy}}
\orcid{0009-0005-3733-2724}

\author{Stefania Montani}
\authornotemark[1]
\affiliation{%
  \institution{Department of Science, Technology and Innovation (DiSIT), Computer Science Institute - Università del Piemonte Orientale, Viale Teresa Michel 11, 15121 Alessandria, Italy}
  \streetaddress{Viale Teresa Michel, 11}
  \streetaddress{Viale Teresa Michel, 11}
  \city{Alessandria}
  \country{Italy}}
\email{stefania.montani@uniupo.it}
\orcid{0000-0002-5992-6735}

\author{Manuel Striani}
\authornote{The authors contributed equally to this research.}
\authornote{Corresponding author. Tel.: +39 0131360179; E-mail address: manuel.striani@uniupo.it}
\affiliation{%
  \institution{Department of Science, Technology and Innovation (DiSIT), Computer Science Institute - Università del Piemonte Orientale, Viale Teresa Michel 11, 15121 Alessandria, Italy}
  \streetaddress{Viale Teresa Michel, 11}
  \city{Alessandria}
  \country{Italy}}
\email{manuel.striani@uniupo.it}
\orcid{0000-0002-7600-576X}

\author{Mariachiara M. Strozzi}
\authornotemark[1]
\affiliation{\institution{Neonatal Intensive Care Unit, Cardinal Massaia Children's Hospital of Asti}\country{Italy}}
\orcid{/0000-0001-8852-7445}


\renewcommand{\shortauthors}{Striani, et al.}

\begin{abstract}

Nowadays, there is evidence that several factors may increase the risk, for an infant, to require stabilisation or resuscitation manoeuvres at birth. However, this risk factors are not completely known, and a universally applicable model for predicting high-risk situations is not available yet. Considering both these limitations and the fact that the need for resuscitation at birth is a rare event, periodic training of the healthcare personnel responsible for newborn caring in the delivery room is mandatory.

In this paper, we propose a machine learning approach for identifying risk factors and their impact on the birth event from real data, which can be used by personnel to progressively increase and update their knowledge.
Our final goal will be the one of designing a user-friendly mobile application, able to improve the recognition rate and the planning of the appropriate interventions on high-risk patients.

\end{abstract}




\maketitle

\section{Introduction}

The transition from fetal to extra-uterine life is characterized by a series of respiratory, cardiovascular and metabolic adaptation mechanisms. Approximately 90\% of newborns breathe spontaneously without the need for interventions, the remaining 10\% will need assistance at birth. Among the latter, most will start breathing after the first assistance maneuvers (drying, tactile stimulation, alignment of the airways); 5\% thanks to the application of positive pressure ventilation (PPV). Estimates of intubation rates vary between 0.4\% and 2\%; less than 0.3\% will require chest compression and approximately 0.05\% will need medication \cite{1,2, 3, 4, 5}.

\begin{figure}
\centering
\includegraphics[width=0.65\textwidth]{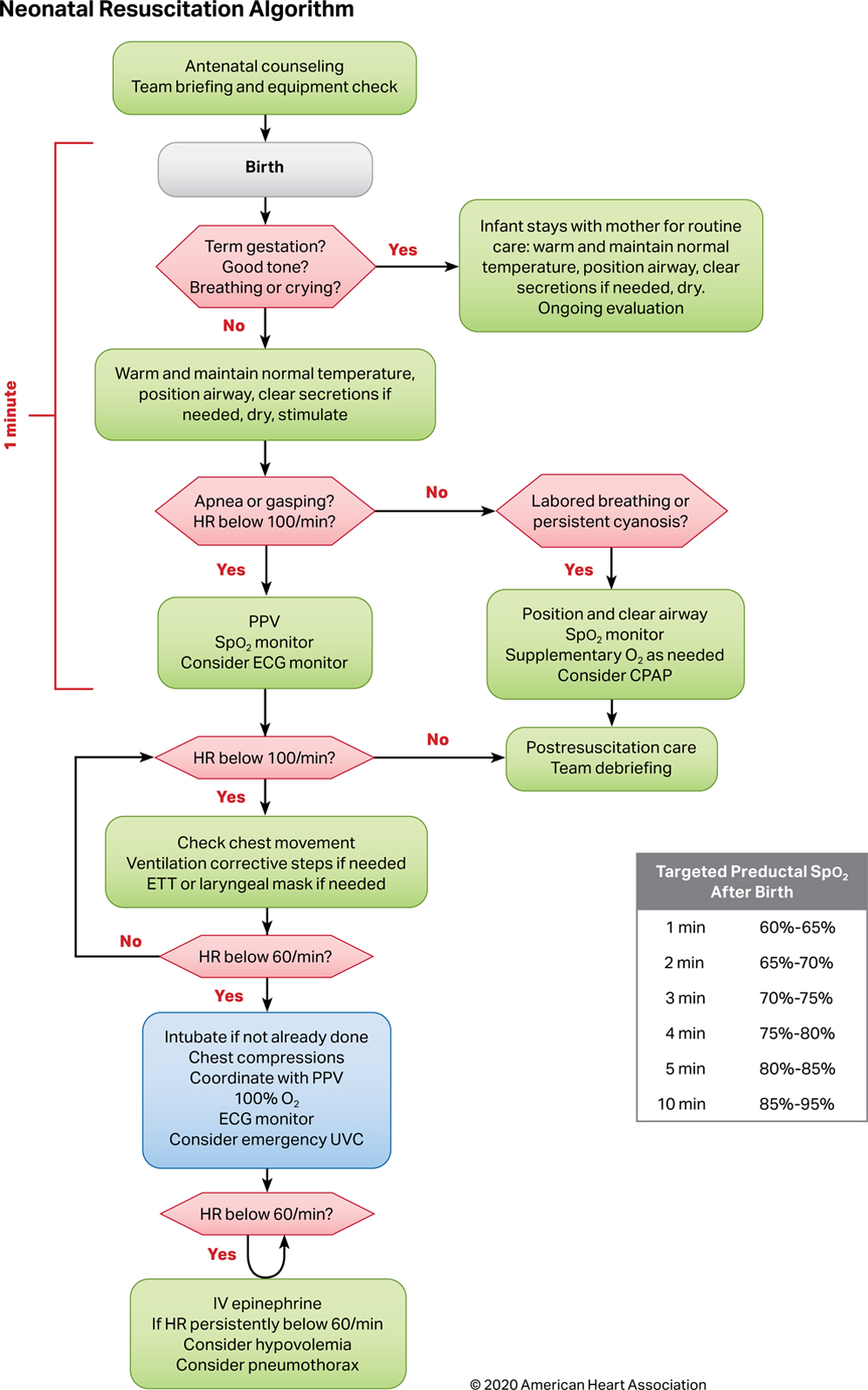}
\caption{Neonatal Resuscitation Algorithm. CPAP indicates continuous positive airway pressure; ECG, electrocardiographic; ETT, endotracheal tube; HR, heart rate; IV, intravenous; $O_2$, oxygen; $SpO_2$, oxygen saturation; and UVC, umbilical venous catheter}. 
\label{NeonatalResuscitationAlgorithm}
\end{figure}

Neonatal mortality in Italy, for babies born after the 22nd week of gestational age, is estimated as 1.7 deaths per 1000 births, compared to an average 2.1/1000 in Europe \cite{6}. The inability of some infants to establish and sustain spontaneous or adequate breathing, contributes significantly to these early deaths and also to the burden of adverse neurological outcomes among survivors. For this reason, it is essential to provide care, ventilation and resuscitation maneuvers to problematic newborns in a timely and effective manner\cite{4,5}. The early identification of high-risk births is the key for timely intervention and  favorable outcome. It is estimated that, in the absence of adequate risk stratification, approximately half of newborns requiring ventilation may not be recognized before birth \cite{4}. However, a general model to predict the resuscitation risk or the need for intervention is not available at the moment, and the list of risk factors reported in the clinical guideline for neonatal rehabilitation (see \ref{NeonatalResuscitationAlgorithm}) is not exhaustive \cite{5, 10}.
Moreover, there is no precise definition of influence degree of these factors, chosen individually or in combination with each other, to define the risk and of assistance at birth and the severity of the newborn conditions.

Considering the rarity of the resuscitation at birth interventions, and the issues mentioned above, it is essential for the healthcare professionals to perform a periodical training, in order to be ready as soon as a newborn caring in the delivery room is needed \cite{4,BOTTRIGHI20232067,StrianiSTCEFMINRTS2023}.

The aim of this project is to evaluate the impact of risk factors on the birth event by means of machine learning techniques, with the final goal of defining a more comprehensive prediction model. The latter will be integrated in a mobile application, to improve the recognition of high-risk patients and to plan appropriate and focused intervention and preparation strategies.

\section{Machine learning for high-risk newborn prediction}

This study includes 250 newborns from the Pediatric Department of the children's hospital "Cardinal Massaia" located in Asti (Italy), from April 2023 to June 2023. No exclusion criteria were applied. The Pediatric Department is a SPOKE center which includes the nursery and neonatal pathology. Patients who require intensive ventilatory/circulatory support or specialist care are stabilized here, with potential support from the Anesthetist, and then sent to the reference HUB center.

The following \textsl{antepartum} and \textsl{intrapartum} risk factors (Tables \ref{MaternalAntepartumRiskFactors}, \ref{AntepartumFetalRiskFactors}, \ref{RiskFactorsIntrapartum}) were collected and stored on an anonymized dataset. We considered only the risk factors considered as important to assess the outcome of the newborn, according to international clinical guidelines \cite {5, 10}.

\begin{table}[t]
\begin{tabular}{@{}l@{}}
\toprule
\textbf{Maternal antepartum risk factors} \\ \midrule
Age $>$ 35 years \\
Previous pathologies or smoking \\
Hypertension \\
Preeclampsia \\
Vaginal swab/rinoculture positive for Streptococcus group B \\
TORCH seroconversion \\
Gestational diabetes \\
Height $<$ 150 cm \\
BMI $>$ 25 \\
Unfollowed pregnancy \\
\bottomrule
\end{tabular}
\caption{Maternal antepartum risk factors}
\label{MaternalAntepartumRiskFactors}
\end{table}
 \begin{table}[t]
        \begin{minipage}{0.5\textwidth}
            \centering
            \begin{tabular}{@{}l@{}}
\toprule
\textbf{Fetal antepartum  risk factors} \\ \midrule
Major malformations \\
IUGR \\
Macrosomia \\
Oligodramnios \\
Polidramnios \\
EG $<$ 37 weeks \\
EG $>$ 41 weeks \\
Birth weight \\
Twin pregnancy \\
Foetal anaemia \\
Lack of steroid prophylaxis in preterm \\
Fetal hydrops \\ \bottomrule
\end{tabular}
\caption{Fetal antepartum  risk factors}
\label{AntepartumFetalRiskFactors}
        \end{minipage}
        \hfillx
        \begin{minipage}{0.5\textwidth}
            \centering
            \begin{tabular}{@{}l@{}}
\toprule
\textbf{Intrapartum risk factors} \\ \midrule
Operative delivery with suction cup/forceps application \\
Vaginal delivery in the breech position \\
Placenta detachment \\
Intrapartum bleeding \\
Prolapse of funiculus / Cord knot \\
Urgent caesarean section \\
Fetal CF pattern type II - III \\
General maternal anaesthesia \\
Shoulder dystocia \\
Chorioamnionitis \\
Amniotic fluid tinged with meconium \\ \bottomrule
\end{tabular}
\caption{Intrapartum risk factors}
\label{RiskFactorsIntrapartum}
        \end{minipage}
    \end{table}

\pagebreak
\newpage

Among the 250 real patients, the following outcomes were registered:

\begin{itemize}
\item 18 were given an APGAR score $\leq 7$ (sampled at 1 minute of life)
\item 12 were ventilated at birth
\item 18 presented respiratory distress after birth
\item 8 were transferred to the NICU (Neonatal Intensive Care Unit)
\item 84 were transferred to Neonatal Pathology
\item Of those who had a brain ultrasound performed: 13 patients had a normal ultrasound, 1 patient had a pathological ultrasound
\item 1 patient underwent passive hypothermia while awaiting STEN
\item 5 required NIV after birth (cPAP, HFNC)
\end{itemize}

We adopted machine learning to predict these outcomes using the risk factors as input, in order to confirm known correlations, as well as to possibly find unknown ones.
Different supervised-machine learning methodologies, namely Decision Tree (DT) and Bayesian Networs (BN) classifiers, were considered \cite{18,19,20}.

The two most frequent outcomes, namely \textbf{APGAR 1} (sampled after 1 minute) and \textbf{Ventilation at birth}, were given particular attention.

\subsection{Results obtained using Decision Trees}

A DT is a predictive model, where each internal node represents a variable, an arc towards a child node represents a possible value for that variable, and a leaf represents the predicted class on the basis of the variable values on the branch. Classifiers based on DTs thus exploit a model that allows the class to be identified by traversing the tree from its root to one of the leaves, based on the values of the observed variables. These classifiers are able to build the tree structure during the training phase, offering a model to be used for inferring the class of new objects provided as queries. 

We trained a DT to predict the APGAR 1 outcome (Figure \ref{BN_APGAR1}). As the acronym suggests, the APGAR score is based on the evaluation of “activity”, “pulse”, “grimace”, “appearance” and “respiration”, and is evaluated at least at 1 and 5 minutes of life for each newborn, assigning a score ranging from 0 to 2 for each item (total score equal to 10). About 90\% of newborns receive an APGAR 1 $ > $ 7, and in these cases no other interventions are usually necessary. In fact, the newborn with an APGAR 1 score in this range presents spontaneous crying, normal skin color and good muscle tone \cite{5, 21}. Otherwise, birth support interventions may be necessary. However, APGAR 1 score alone cannot be considered an index of asphyxia, and it does not predict the possibility of neurological damage or perinatal death \cite{21}.

\begin{figure}[t]
\centering
\includegraphics[width=0.5\textwidth]{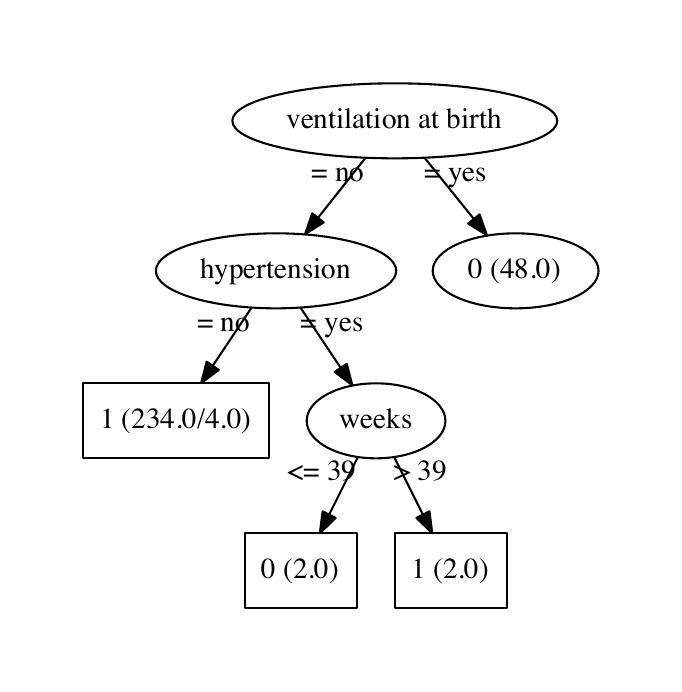}
\caption{
Decision Tree obtained using \textbf{APGAR 1} as a class by increasing the sample with the SMOTE technique ($accuracy=97.2\%$)}. 
\label{VentNascita_YesSMOTE}
\end{figure}

Due to the low number of "pathological" newborns  with APGAR 1 $\leq 7$ (class 0), the dataset was heavily unbalanced. Therefore, we used a statistical approach called SMOTE \cite{Chawla2002SMOTESM}, to balance the rate between overrepresented and underrepresented cases. After the adoption of this technique, the number of patients with APGAR 1 $\leq 7$, initially 18, was raised to 54. The DT obtained with the latter dataset is shown in Figure \ref{VentNascita_YesSMOTE}. 

The leaves of the tree represent the classes, i.e. the values of APGAR (class 0: APGAR 1 $\leq7$, class 1: APGAR 1 $\geq 8$), while the other (internal) nodes represent the variables that contribute to the assignment of the class, on the basis of their values.

As it can be observed from the Figure, three  variables were identified as relevant for prediction:
\begin{itemize}
\item Ventilation at birth
\item Maternal hypertension
\item EG $< 39$ weeks
\end{itemize}

This finding is coherent with the medical knowledge. In particular, 
there is a correlation between patients with a low APGAR score at birth and the probability that they will be ventilated on the neonatal care device.
Concerning maternal hypertension, it is known  that those born to mothers with non-severe/well-controlled chronic hypertension do not appear to be at risk of complications \cite{22}. However, when the mother suffers from severe hypertension, the risk of premature birth, IUGR, placental abruption appears to be increased compared to the general population \cite{23}. 
Lastly, it is certainly known that preterm newborns often need help in the transition from fetal to extra-uterine life.  

Our model reached an accuracy of 97.2\%. The results are shown in Tables \ref{results_table_j48_smote} and \ref{ConfusionMatrixDT_J48_YesSMOTE}.


\begin{table}[ht]
\begin{tabular}{rccccccccc}
\hline
\textbf{Class} & \textbf{TP Rate} & \textbf{FP Rate} & \textbf{Precision} & \textbf{Recall} & \textbf{F-Measure} & \textbf{MCC} & \textbf{ROC Area} & \textbf{PRC Area} & \textbf{Accuracy} \\ \hline
0 (APGAR 1 $\leq 7 $)& 0.89 & 0.01 & 0.96 & 0.889 & 0.923 & 0.91 & 0.92 & 0.91 &  \\
1 (APGAR 1 $\geq 8$) & 0.99 & 0.11 & 0.98 & 0.991 & 0.983 & 0.91 & 0.92 & 0.96 &  \\
\textbf{Weighted Avg.} & 0.97 & 0.09 & 0.97 & 0.97 & 0.97 & 0.91 & 0.92 & 0.95 & 0.972 \\ \hline
\end{tabular}
\caption{Results obtained by the Decision Tree classifier with the SMOTE technique}
\label{results_table_j48_smote}
\end{table}

\begin{table}[ht]
\begin{tabular}{@{}ccl@{}}
\toprule
a & b & \textless{}-- classified as \\ \midrule
48 & 6 & $|$ a = 0 \\
223 & 0 & $|$  b = 1 \\ \bottomrule
\end{tabular}
\caption{Confusion matrix for Decision Tree classifier with the SMOTE technique}
\label{ConfusionMatrixDT_J48_YesSMOTE}
\end{table}

\subsection{Results obtained using Bayesian Networks}

A Bayesian Network (BN) is a probabilistic model that allows the graphical representation of the statistical dependence relationships between the variables of a system. Bayesian classifiers are based on a BN model, which describes the probabilistic interrelationships between attributes and classes. During the training phase, the structure and probability distribution of the network are calculated in order to maximize the relationships between the data and the corresponding classes. Then, given a new item, the classifier applies the value of its attributes as observations in the network and then assigns the resulting class with the highest probability.

\begin{figure}[H]
\centering 
\includegraphics[width=\textwidth]{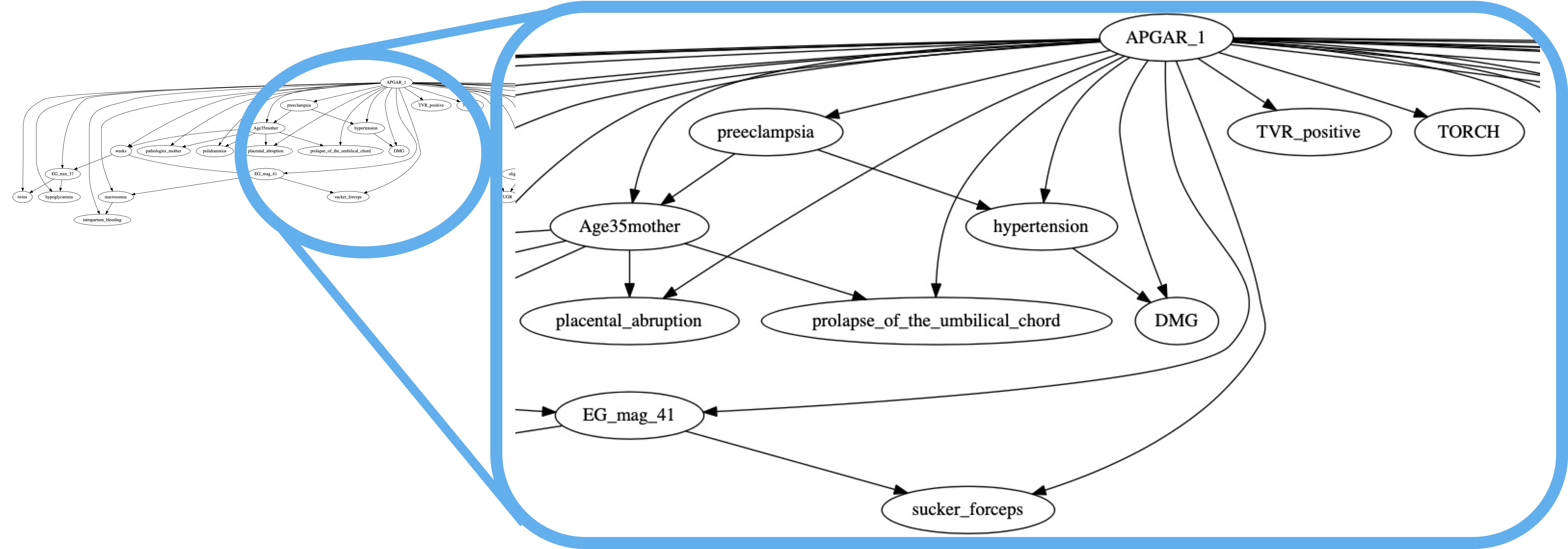} 
\caption{Bayesian network obtained by using \textsl{APGAR 1} as a class ($accuracy=90.2\%$)}. 
\label{BN_APGAR1}
\end{figure}

In our application, the use of a BN allowed to recognize which variables statistically influence others when predicting the class (outcome APGAR 1 $ \leq 7$), as graphically represented in Figure \ref{BN_APGAR1}.


The accuracy of this model was found to be 90.2\%. As can be seen from Figure \ref{BN_APGAR1}, the correlations/causations found are much more numerous than those found using the DT, but always  well justified by medical knowledge.

\begin{figure}
\centering 
\includegraphics[width=\textwidth]{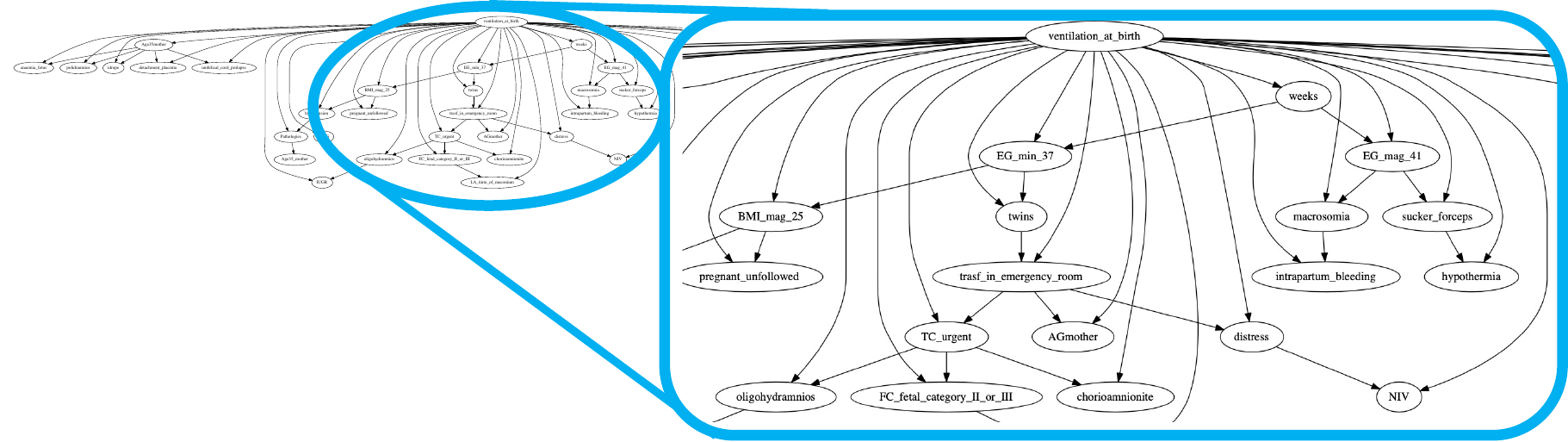}
\caption{Bayesian network obtained by using \textsl{Ventilation at birth} as a class ($accuracy=89.16\%$)}. 
\label{BN_VentNascita}
\end{figure}

We also learned a BN classifier predicting the outcome "ventilation at birth" as a class (see Figure \ref{BN_VentNascita}), obtaining findings which are coherent with the literature as well. 


It is important to note that, in Figure \ref{BN_APGAR1} and Figure \ref{BN_VentNascita}, inter-connections and correations were also found between the different risk factors. These correlations are:
\begin{itemize}
\item EG $>$ 41 weeks, with macrosomia and ventouse/forceps (variables \textsl{EG\_mag\_41}, \textsl{macrosomia} and \textsl{sucker\_forceps})
\item EG $\leq 37$ weeks, with twin pregnancy (variables \textsl{EG\_min\_37} and \textsl{twins})
\item Maternal age $>$ 35 years with maternal pathologies (like hypertension) and  transfer to emergency room (variables \textsl{AGmother}, \textsl{distress} and \textsl{trasf\_in\_emergency\_room})
\item Urgent CT, associated with category II or III CF, oligohydramnios, chorioamnionitis (with IUGR) (variables \textsl{TC\_urgent}, \textsl{FC\_fetal\_category\_II\_or\_III}, \textsl{chorioamnionitis} and \textsl{oligohydramnios})
    \item EG $\leq$ 37 weeks, which in turn correlates with twins and maternal BMI $>$ 25 (variables \textsl{EG\_min\_37}, \textsl{twins} and \textsl{BMI\_mag\_25})
    \item Respiratory distress and NIV (Non-Invasive Ventilation) (variables \textsl{distress} and \textsl{NIV})
    \item Macrosomia, related to intrapartum bleeding and EG $>$ 41 weeks(variables \textsl{macrosomia}, \textsl{intrapartum\_bleeding} and \textsl{EG\_mag\_41}) 
\end{itemize}

These findings are meaningful. The correlation between twin pregnancy and prematurity is well known.
The correlation between oligohydramnios and IUGR is also known as normal fetal growth is determined by genetic heritage, maternal factors and placental "health". Intrauterine growth restriction occurs when the genetic potential of the fetus is not reached due to alterations in one or more of these factors. Endo-uterine growth restriction determines a risk factor for the development of oligohydramnios, which occurs when the balance between the factors that cause an increase in amniotic fluid (fetal diuresis, pulmonary secretion) and factors that cause its reduction (swallowing , reabsorption at the level of the uterine mucosa), is unbalanced towards the latter \cite{25}.
The association present in the literature between advanced maternal age and placental abruption and the increased incidence of maternal pathologies, which could have consequences on fetal well-being, was also confirmed \cite{17}.

In conclusion, the machine learning predictors we adopted were able to confirm known correlations and causations between the variables at play, and made them explicit within more comprehensive models; in particular, the BN model was particularly rich, and was able to highlight inter-connections between the variables which are less trivial.

Interestingly, both Bayesian classifiers and Decision Trees also provide a graphical view of the classification model, which improves explainability.

Such models can therefore be proposed to the personnel of the birth room with educational purposes, given their simplicity and interpretability.

\begin{figure}[H]
\centering
\includegraphics[scale=0.35]{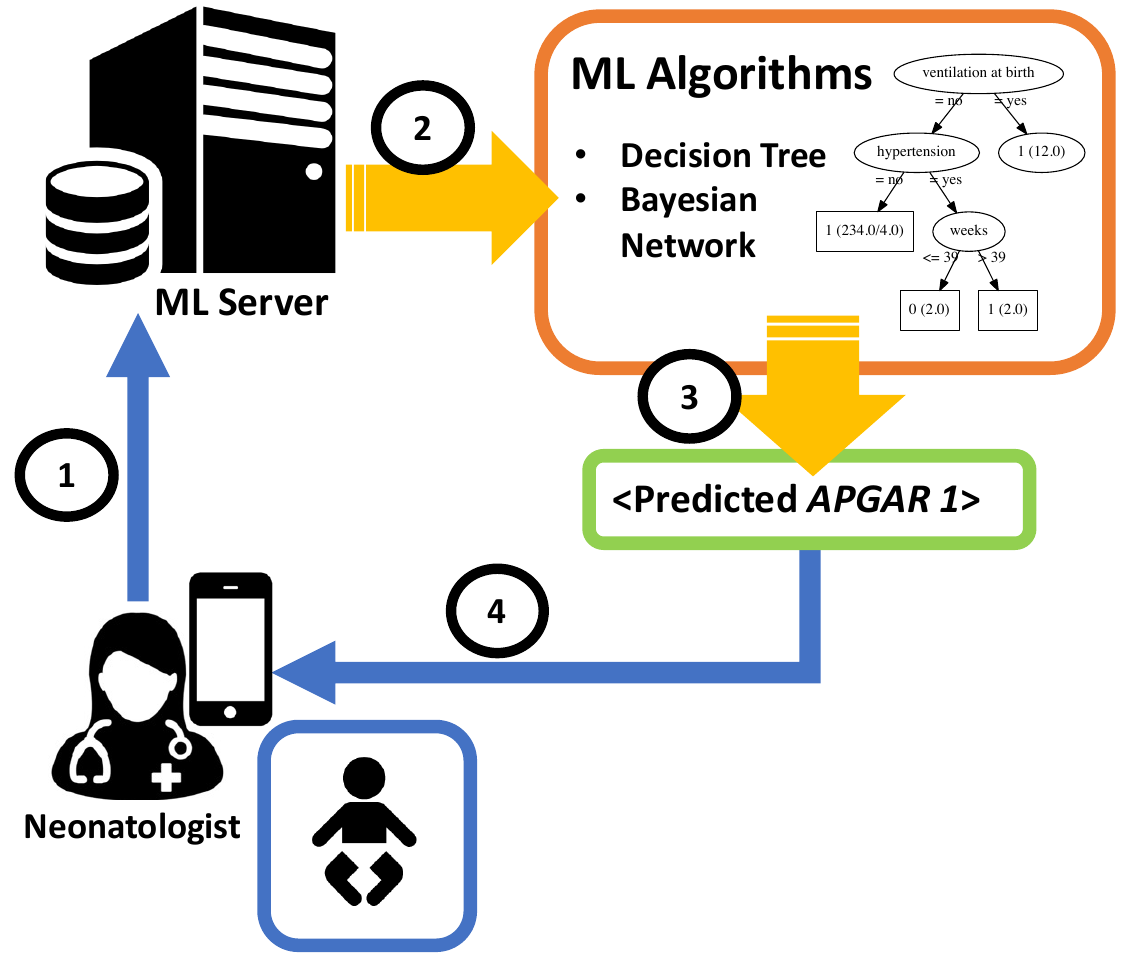}
\caption{The workflow of the educational system}
\label{workflowArchitecture}
\end{figure}

Figure \ref{workflowArchitecture} shows the workflow of our educational tool. In particular, the neonatologists, by using a mobile app, through a minimalist interface, can insert some risk factors and sent a query to a server (steps 1 and 2 in the Figure). The server applies
the prediction models, and sends the results (e.g., predicted APGAR 1 score) to the mobile app, easily usable in the delivery room (steps 3 and 4).

This educational tool is under development at our Computer Science department.

\section{Discussion and conclusion}

Thanks to the supervised machine learning techniques used, it was possible to learn models, such as Decision Trees and Bayesian Networks, able to classify the patients in our the dataset and to highlight correlations between intrapartum and antepartum factors and the consequent neonatal outcomes. These correlations confirm some of the findings described in the literature, making them explicit in easy to read comprehensive models.

Our models show that adding interplaying risk factors increases the probability of unfavorable outcomes. This is a non trivial result, that was provided by the analysis of the BN models. Such an outcome could allow for a better recognition of risky situations and can be considered as an added value in the clinical practice.


These first results testify the applicability of the approach to, on one side, improve the treatment strategies for problematic newborns and, on the other side, to tailor the educational tools for educational purposes, suggesting the trainees the more appropriate procedures and pre-alert strategies when a risky situation is about to happen. In order to further improve these results, a broader collection of cases should be provided to train more informative machine learning models. These models will be integrated, as a final step, into a user-friendly mobile application, tailored to the needs of the resuscitation personnel.


\bibliographystyle{ACM-Reference-Format}
\bibliography{sample-base}


\begin{thebibliography}{18}


\ifx \showCODEN    \undefined \def \showCODEN     #1{\unskip}     \fi
\ifx \showDOI      \undefined \def \showDOI       #1{#1}\fi
\ifx \showISBNx    \undefined \def \showISBNx     #1{\unskip}     \fi
\ifx \showISBNxiii \undefined \def \showISBNxiii  #1{\unskip}     \fi
\ifx \showISSN     \undefined \def \showISSN      #1{\unskip}     \fi
\ifx \showLCCN     \undefined \def \showLCCN      #1{\unskip}     \fi
\ifx \shownote     \undefined \def \shownote      #1{#1}          \fi
\ifx \showarticletitle \undefined \def \showarticletitle #1{#1}   \fi
\ifx \showURL      \undefined \def \showURL       {\relax}        \fi
\providecommand\bibfield[2]{#2}
\providecommand\bibinfo[2]{#2}
\providecommand\natexlab[1]{#1}
\providecommand\showeprint[2][]{arXiv:#2}

\bibitem[2021({[n.\,d.]})]%
        {5}
\bibfield{author}{\bibinfo{person}{ERC~Guidelines 2021}.}
  \bibinfo{year}{[n.\,d.]}\natexlab{}.
\newblock
\newblock
\urldef\tempurl%
\url{https://www.ircouncil.it/linee-guida-rcp-2021/}
\showURL{%
\tempurl}


\bibitem[Anna-Karin~Wikström et~al\mbox{.}(2016)]%
        {22}
\bibfield{author}{\bibinfo{person}{Johanna~Gunnarsdottir Anna-Karin~Wikström},
  \bibinfo{person}{Maria Nelander}, \bibinfo{person}{Marija Simic},
  \bibinfo{person}{Olof Stephansson}, {and} \bibinfo{person}{Sven
  Cnattingius}.} \bibinfo{year}{2016}\natexlab{}.
\newblock \showarticletitle{Prehypertension in Pregnancy and Risks of Small for
  Gestational Age Infant and Stillbirth}.
\newblock \bibinfo{journal}{\emph{Hypertension}} \bibinfo{volume}{67},
  \bibinfo{number}{3} (\bibinfo{year}{2016}), \bibinfo{pages}{640--646}.
\newblock
\urldef\tempurl%
\url{https://doi.org/10.1161/HYPERTENSIONAHA.115.06752}
\showDOI{\tempurl}


\bibitem[Aziz et~al\mbox{.}(2008)]%
        {10}
\bibfield{author}{\bibinfo{person}{Khalid Aziz}, \bibinfo{person}{Mairi
  Chadwick}, \bibinfo{person}{Mary Baker}, {and} \bibinfo{person}{Wayne
  Andrews}.} \bibinfo{year}{2008}\natexlab{}.
\newblock \showarticletitle{Ante- and intra-partum factors that predict
  increased need for neonatal resuscitation}.
\newblock \bibinfo{journal}{\emph{Resuscitation}} \bibinfo{volume}{79},
  \bibinfo{number}{3} (\bibinfo{year}{2008}), \bibinfo{pages}{444--452}.
\newblock
\showISSN{0300-9572}
\urldef\tempurl%
\url{https://doi.org/10.1016/j.resuscitation.2008.08.004}
\showDOI{\tempurl}


\bibitem[Aziz et~al\mbox{.}(2021)]%
        {4}
\bibfield{author}{\bibinfo{person}{Khalid Aziz}, \bibinfo{person}{Chair;
  Henry~C. Lee}, \bibinfo{person}{Marilyn~B. Escobedo},
  \bibinfo{person}{Amber~V. Hoover}, \bibinfo{person}{Beena~D. Kamath-Rayne},
  \bibinfo{person}{Vishal~S. Kapadia}, \bibinfo{person}{David~J. Magid},
  \bibinfo{person}{Susan Niermeyer}, \bibinfo{person}{Georg~M. Schmölzer},
  \bibinfo{person}{Edgardo Szyld}, \bibinfo{person}{Gary~M. Weiner},
  \bibinfo{person}{Myra~H. Wyckoff}, \bibinfo{person}{Nicole~K. Yamada}, {and}
  \bibinfo{person}{Jeanette Zaichkin}.} \bibinfo{year}{2021}\natexlab{}.
\newblock \showarticletitle{Part 5: Neonatal Resuscitation 2020 American Heart
  Association Guidelines for Cardiopulmonary Resuscitation and Emergency
  Cardiovascular Care}.
\newblock \bibinfo{journal}{\emph{Pediatrics}} \bibinfo{volume}{147},
  \bibinfo{number}{Supplement 1} (\bibinfo{date}{01} \bibinfo{year}{2021}),
  \bibinfo{pages}{e2020038505E}.
\newblock
\showISSN{0031-4005}
\urldef\tempurl%
\url{https://doi.org/10.1542/peds.2020-038505E}
\showDOI{\tempurl}


\bibitem[Berazategui et~al\mbox{.}(2017)]%
        {2}
\bibfield{author}{\bibinfo{person}{Juan~Pablo Berazategui},
  \bibinfo{person}{Adriana Aguilar}, \bibinfo{person}{Marilyn Escobedo},
  \bibinfo{person}{Douglas Dannaway}, \bibinfo{person}{Ruth Guinsburg},
  \bibinfo{person}{Maria Fernanda~Branco de Almeida}, \bibinfo{person}{Firas
  Saker}, \bibinfo{person}{Ariel Fern{\'a}ndez}, \bibinfo{person}{Guadalupe
  Albornoz}, \bibinfo{person}{Mariana Valera}, \bibinfo{person}{Daniel Amado},
  \bibinfo{person}{Gabriela Puig}, \bibinfo{person}{Fernando Althabe}, {and}
  \bibinfo{person}{Edgardo Szyld}.} \bibinfo{year}{2017}\natexlab{}.
\newblock \showarticletitle{Risk factors for advanced resuscitation in term and
  near-term infants: a case{\textendash}control study}.
\newblock \bibinfo{journal}{\emph{Archives of Disease in Childhood - Fetal and
  Neonatal Edition}} \bibinfo{volume}{102}, \bibinfo{number}{1}
  (\bibinfo{year}{2017}), \bibinfo{pages}{F44--F50}.
\newblock
\showISSN{1359-2998}
\urldef\tempurl%
\url{https://doi.org/10.1136/archdischild-2015-309525}
\showDOI{\tempurl}
\showeprint{https://fn.bmj.com/content/102/1/F44.full.pdf}


\bibitem[Bottrighi et~al\mbox{.}(2023)]%
        {BOTTRIGHI20232067}
\bibfield{author}{\bibinfo{person}{Alessio Bottrighi}, \bibinfo{person}{Marco
  Guazzone}, \bibinfo{person}{Giorgio Leonardi}, \bibinfo{person}{Stefania
  Montani}, \bibinfo{person}{Manuel Striani}, {and} \bibinfo{person}{Paolo
  Terenziani}.} \bibinfo{year}{2023}\natexlab{}.
\newblock \showarticletitle{Applying the SIM Tool in Clinical Practice: a Case
  Study in Neonatal Resuscitation Simulation}.
\newblock \bibinfo{journal}{\emph{Procedia Computer Science}}
  \bibinfo{volume}{225} (\bibinfo{year}{2023}), \bibinfo{pages}{2067--2075}.
\newblock
\showISSN{1877-0509}
\urldef\tempurl%
\url{https://doi.org/10.1016/j.procs.2023.10.197}
\showDOI{\tempurl}
\newblock
\shownote{27th International Conference on Knowledge Based and Intelligent
  Information and Engineering Sytems (KES 2023)}.


\bibitem[Canonico et~al\mbox{.}(2023)]%
        {StrianiSTCEFMINRTS2023}
\bibfield{author}{\bibinfo{person}{Massimo Canonico}, \bibinfo{person}{Stefania
  Montani}, {and} \bibinfo{person}{Manuel Striani}.}
  \bibinfo{year}{2023}\natexlab{}.
\newblock \showarticletitle{NRTS: A Client-Server Architecture for Supporting
  Education in a Neonatal Resuscitation Simulation Scenario}.
\newblock \bibinfo{journal}{\emph{Studies in health technology and
  informatics}}  \bibinfo{volume}{309} (\bibinfo{year}{2023}),
  \bibinfo{pages}{97--98}.
\newblock
\urldef\tempurl%
\url{https://doi.org/10.3233/SHTI230748}
\showDOI{\tempurl}
\newblock
\shownote{Telehealth Ecosystems in Practice}.


\bibitem[Chawla et~al\mbox{.}(2002)]%
        {Chawla2002SMOTESM}
\bibfield{author}{\bibinfo{person}{Nitesh~V Chawla}, \bibinfo{person}{Kevin~W
  Bowyer}, \bibinfo{person}{Lawrence~O Hall}, {and} \bibinfo{person}{W~Philip
  Kegelmeyer}.} \bibinfo{year}{2002}\natexlab{}.
\newblock \showarticletitle{SMOTE: synthetic minority over-sampling technique}.
\newblock \bibinfo{journal}{\emph{Journal of artificial intelligence research}}
   \bibinfo{volume}{abs/1106.1813} (\bibinfo{year}{2002}),
  \bibinfo{pages}{321--357}.
\newblock
\urldef\tempurl%
\url{https://api.semanticscholar.org/CorpusID:1554582}
\showURL{%
\tempurl}


\bibitem[FETUS et~al\mbox{.}(2015)]%
        {21}
\bibfield{author}{\bibinfo{person}{AMERICAN ACADEMY OF PEDIATRICS COMMITTEE~ON
  FETUS}, \bibinfo{person}{NEWBORN}, \bibinfo{person}{AMERICAN COLLEGE~OF
  OBSTETRICIANS}, \bibinfo{person}{GYNECOLOGISTS COMMITTEE ON~OBSTETRIC
  PRACTICE}, \bibinfo{person}{Kristi~L. Watterberg}, \bibinfo{person}{Susan
  Aucott}, \bibinfo{person}{William~E. Benitz}, \bibinfo{person}{James~J.
  Cummings}, \bibinfo{person}{Eric~C. Eichenwald}, \bibinfo{person}{Jay
  Goldsmith}, \bibinfo{person}{Brenda~B. Poindexter}, \bibinfo{person}{Karen
  Puopolo}, \bibinfo{person}{Dan~L. Stewart}, \bibinfo{person}{Kasper~S. Wang},
  \bibinfo{person}{Jeffrey~L. Ecker}, \bibinfo{person}{Joseph~R. Wax},
  \bibinfo{person}{Ann Elizabeth~Bryant Borders}, \bibinfo{person}{Yasser~Yehia
  El-Sayed}, \bibinfo{person}{R.~Phillips Heine}, \bibinfo{person}{Denise~J.
  Jamieson}, \bibinfo{person}{Maria~Anne Mascola}, \bibinfo{person}{Howard~L.
  Minkoff}, \bibinfo{person}{Alison~M. Stuebe}, \bibinfo{person}{James~E.
  Sumners}, \bibinfo{person}{Methodius~G. Tuuli}, {and}
  \bibinfo{person}{Kurt~R. Wharton}.} \bibinfo{year}{2015}\natexlab{}.
\newblock \showarticletitle{{The Apgar Score}}.
\newblock \bibinfo{journal}{\emph{Pediatrics}} \bibinfo{volume}{136},
  \bibinfo{number}{4} (\bibinfo{date}{10} \bibinfo{year}{2015}),
  \bibinfo{pages}{819--822}.
\newblock
\showISSN{0031-4005}
\urldef\tempurl%
\url{https://doi.org/10.1542/peds.2015-2651}
\showDOI{\tempurl}
\showeprint{https://publications.aap.org/pediatrics/article-pdf/136/4/819/1060316/peds\_2015-2651.pdf}


\bibitem[Kahveci et~al\mbox{.}(2018)]%
        {17}
\bibfield{author}{\bibinfo{person}{Bekir Kahveci}, \bibinfo{person}{Rauf
  Melekoglu}, \bibinfo{person}{Ismail~Cuneyt Evruke}, {and}
  \bibinfo{person}{Cihan Cetin}.} \bibinfo{year}{2018}\natexlab{}.
\newblock \showarticletitle{The effect of advanced maternal age on perinatal
  outcomes in nulliparous singleton pregnancies}.
\newblock \bibinfo{journal}{\emph{BMC Pregnancy and Childbirth}}
  \bibinfo{volume}{18}, \bibinfo{number}{1} (\bibinfo{date}{Aug.}
  \bibinfo{year}{2018}), \bibinfo{pages}{343}.
\newblock
\showISSN{1471-2393}
\urldef\tempurl%
\url{https://doi.org/10.1186/s12884-018-1984-x}
\showDOI{\tempurl}


\bibitem[perinatale. V~European Perinatal Health Report~(iss.it)({[n.\,d.]})]%
        {6}
\bibfield{author}{\bibinfo{person}{Salute perinatale. V~European Perinatal
  Health Report~(iss.it)}.} \bibinfo{year}{[n.\,d.]}\natexlab{}.
\newblock
\newblock
\urldef\tempurl%
\url{https://www.epicentro.iss.it/materno/report-europeristat-2022}
\showURL{%
\tempurl}


\bibitem[Quinlan(1986)]%
        {20}
\bibfield{author}{\bibinfo{person}{J.~R. Quinlan}.}
  \bibinfo{year}{1986}\natexlab{}.
\newblock \showarticletitle{Induction of decision trees}.
\newblock \bibinfo{journal}{\emph{Machine Learning}} \bibinfo{volume}{1},
  \bibinfo{number}{1} (\bibinfo{date}{March} \bibinfo{year}{1986}),
  \bibinfo{pages}{81--106}.
\newblock
\showISSN{1573-0565}
\urldef\tempurl%
\url{https://doi.org/10.1007/BF00116251}
\showDOI{\tempurl}


\bibitem[Ross et~al\mbox{.}(2008)]%
        {25}
\bibfield{author}{\bibinfo{person}{{Michael G.} Ross}, \bibinfo{person}{Ron
  Beloosesky}, {and} \bibinfo{person}{{John T.} Queenan}.}
  \bibinfo{year}{2008}\natexlab{}.
\newblock \bibinfo{booktitle}{\emph{Polyhydramnios and Oligohydramnios}}.
\newblock \bibinfo{pages}{316--325}.
\newblock
\showISBNx{1405127821}
\urldef\tempurl%
\url{https://doi.org/10.1002/9780470691878.ch38}
\showDOI{\tempurl}


\bibitem[Shalev-Shwartz and Ben-David(2014)]%
        {19}
\bibfield{author}{\bibinfo{person}{Shai Shalev-Shwartz} {and}
  \bibinfo{person}{Shai Ben-David}.} \bibinfo{year}{2014}\natexlab{}.
\newblock \bibinfo{booktitle}{\emph{Understanding Machine Learning: From Theory
  to Algorithms}}.
\newblock \bibinfo{publisher}{Cambridge University Press},
  \bibinfo{address}{USA}.
\newblock
\showISBNx{1107057132}


\bibitem[Szyld et~al\mbox{.}(2014)]%
        {3}
\bibfield{author}{\bibinfo{person}{Edgardo Szyld}, \bibinfo{person}{Adriana
  Aguilar}, \bibinfo{person}{Gabriel~A. Musante}, \bibinfo{person}{Nestor
  Vain}, \bibinfo{person}{Luis Prudent}, \bibinfo{person}{Jorge Fabres}, {and}
  \bibinfo{person}{Waldemar~A. Carlo}.} \bibinfo{year}{2014}\natexlab{}.
\newblock \showarticletitle{Comparison of Devices for Newborn Ventilation in
  the Delivery Room}.
\newblock \bibinfo{journal}{\emph{The Journal of Pediatrics}}
  \bibinfo{volume}{165}, \bibinfo{number}{2} (\bibinfo{year}{2014}),
  \bibinfo{pages}{234--239.e3}.
\newblock
\showISSN{0022-3476}
\urldef\tempurl%
\url{https://doi.org/10.1016/j.jpeds.2014.02.035}
\showDOI{\tempurl}


\bibitem[Szyld et~al\mbox{.}(2022)]%
        {1}
\bibfield{author}{\bibinfo{person}{Edgardo Szyld}, \bibinfo{person}{Michael~P
  Anderson}, \bibinfo{person}{Birju~A Shah}, \bibinfo{person}{Charles~C Roehr},
  \bibinfo{person}{Georg~M Schm{\"o}lzer}, \bibinfo{person}{Jorge~G Fabres},
  {and} \bibinfo{person}{Gary~M Weiner}.} \bibinfo{year}{2022}\natexlab{}.
\newblock \showarticletitle{Risk calculator for advanced neonatal
  resuscitation}.
\newblock \bibinfo{journal}{\emph{BMJ Paediatrics Open}} \bibinfo{volume}{6},
  \bibinfo{number}{1} (\bibinfo{year}{2022}).
\newblock
\urldef\tempurl%
\url{https://doi.org/10.1136/bmjpo-2021-001376}
\showDOI{\tempurl}
\showeprint{https://bmjpaedsopen.bmj.com/content/6/1/e001376.full.pdf}


\bibitem[Villar et~al\mbox{.}(2006)]%
        {23}
\bibfield{author}{\bibinfo{person}{José Villar}, \bibinfo{person}{Guillermo
  Carroli}, \bibinfo{person}{Daniel Wojdyla}, \bibinfo{person}{Edgardo Abalos},
  \bibinfo{person}{Daniel Giordano}, \bibinfo{person}{Hassan Ba'aqeel},
  \bibinfo{person}{Ubaldo Farnot}, \bibinfo{person}{Per Bergsjø},
  \bibinfo{person}{Leiv Bakketeig}, \bibinfo{person}{Pisake Lumbiganon},
  \bibinfo{person}{Liana Campodónico}, \bibinfo{person}{Yagob Al-Mazrou},
  \bibinfo{person}{Marshall Lindheimer}, {and} \bibinfo{person}{Michael
  Kramer}.} \bibinfo{year}{2006}\natexlab{}.
\newblock \showarticletitle{Preeclampsia, gestational hypertension and
  intrauterine growth restriction, related or independent conditions?}
\newblock \bibinfo{journal}{\emph{American Journal of Obstetrics and
  Gynecology}} \bibinfo{volume}{194}, \bibinfo{number}{4}
  (\bibinfo{year}{2006}), \bibinfo{pages}{921--931}.
\newblock
\showISSN{0002-9378}
\urldef\tempurl%
\url{https://doi.org/10.1016/j.ajog.2005.10.813}
\showDOI{\tempurl}


\bibitem[Witten et~al\mbox{.}(2011)]%
        {18}
\bibfield{editor}{\bibinfo{person}{Ian~H. Witten}, \bibinfo{person}{Eibe
  Frank}, {and} \bibinfo{person}{Mark~A. Hall}} (Eds.).
  \bibinfo{year}{2011}\natexlab{}.
\newblock \bibinfo{booktitle}{ (\bibinfo{edition}{third edition} ed.)}.
\newblock \bibinfo{publisher}{Morgan Kaufmann}, \bibinfo{address}{Boston}.
  587--605 pages.
\newblock
\showISBNx{978-0-12-374856-0}
\urldef\tempurl%
\url{https://doi.org/10.1016/B978-0-12-374856-0.00023-7}
\showDOI{\tempurl}


\end{thebibliography}

\end{document}